\documentclass[runningheads]{llncs}
\usepackage{graphicx}
\usepackage{subcaption}
\begin{document}
\title{Liquid Leak Detection Using Thermal Images}
\author{Kalpak Bansod\and
        Yanshan Wan \and
        Yugesh Rai}
\authorrunning{Kalpak Bansod, Yanshan Wan, Yugesh Rai}
\institute{Department of Computing Science - University of Alberta \\
\email{kalpakni@ualberta.ca, yanshan@ualberta.ca, yugesh@ualberta.ca}}
\maketitle

\begin{abstract}
This paper presents a comprehensive solution to address the critical challenge of liquid leaks in the oil and gas industry, leveraging advanced computer vision and deep learning methodologies. Employing You Only Look Once (YOLO) and Real-Time Detection Transformer (RT DETR) models, our project focuses on enhancing early identification of liquid leaks in key infrastructure components such as pipelines, pumps, and tanks. Through the integration of surveillance thermal cameras and sensors, the combined YOLO and RT DETR models demonstrate remarkable efficacy in the continuous monitoring and analysis of visual data within oil and gas facilities. YOLO's real-time object detection capabilities swiftly recognize leaks and their patterns, while RT DETR excels in discerning specific leak-related features, particularly in thermal images. This approach significantly improves the accuracy and speed of leak detection, ultimately mitigating environmental and financial risks associated with liquid leaks.

\end{abstract}

\section{Introduction}
The oil and gas industry faces a critical challenge in preventing and mitigating the impact of liquid leaks, which can lead to severe environmental damage and substantial financial losses. This paper addresses this issue through the implementation of advanced computer vision and deep learning techniques. By incorporating the You Only Look Once (YOLO) and Real-Time Detection Transformer (RT DETR) models, we aim to implement the identification of liquid leaks in key infrastructure components. Thermal images form the backbone of our monitoring system, enabling the YOLO and RT DETR models to analyze visual data despite weather and time of day. YOLO's real-time object detection capabilities enable swift recognition of leaks and associated patterns, while RT DETR excels in identifying subtle features, particularly in thermal images.

This paper provides a detailed account of the training process, where deep neural networks are trained to identify patterns and irregularities associated with liquid leaks, facilitating real-time detection. The integration of YOLO and RT DETR not only distinguishes liquid leaks from normal operations but also significantly reduces response times, thus preventing minor leaks from escalating into environmentally and financially burdensome disasters. A comparative analysis highlights the complementary strengths of YOLO and RT DETR, with the integrated system ensuring a robust and accurate solution for early leak detection. In conclusion, this project represents a significant leap forward in utilizing cutting-edge technology to enhance the safety and sustainability of the oil and gas industry, offering a forward-looking solution for environmental and financial risk mitigation.

\subsection{Literature Review}

\subsubsection{End-to-End Object Detection with Transformers \cite{carion2020end}}
In ‘End-to-End Object Detection with Transformers’, Carion and colleagues (2020) noted that DETR is a new design for object detection systems based on transformers and bipartite matching loss for direct set prediction. It achieves competitive results compared to Faster R-CNN in quantitative evaluation on COCO. DETR can also be generalized to produce panoptic segmentation in a unified manner and outperforms competitive baselines. The architecture of DETR is versatile and extensible, and it performs significantly better on large objects. DETR demonstrates accuracy and run-time performance on par with the well-established and highly-optimized Faster RCNN baseline on the challenging COCO object detection dataset. Moreover, DETR can be easily generalized to produce panoptic segmentation in a unified manner. The authors show that it significantly outperforms competitive baselines. However, there are challenges in training, optimization, and performance on small objects.
Aspects of the researchers’ results potentially reinforce earlier work in this area: “DETR does not require NMS due to its set-based loss,” Carion suggested.
They advocate that the new design for detectors presents challenges in training, optimization, and performance on small objects. However, with future work, these challenges can be successfully addressed for DETR.

\subsubsection{DETRs Beat YOLOs on Real-time Object Detection \cite{lv2023detrs}}
The paper discusses the challenges associated with the high computational cost of DETRs (Data-efficient Transformers for Object Detection) and the limitations it imposes on practical applications of DETRs in real-time object detection scenarios. To address this issue, the paper introduces a novel approach that focuses on improving the efficiency of the encoder in processing multi-scale features and enabling flexible adjustment of inference speed without retraining the model.
The research emphasizes the importance of selecting encoder features using IoU-aware query selection, highlighting that this not only increases the proportion of high classification scores but also results in more features with high classification scores and high IoU scores. The evaluation on val2017 demonstrates an improvement of 0.8 percent Average Precision (AP) when employing IoU-aware query selection. Additionally, the paper discusses the impact of Non-Maximum Suppression (NMS) on the inference speed of real-time detectors, establishing the need for more efficient mechanisms to improve real-time object detection performance.

\subsubsection{YOLO-IMF: An Improved YOLOv8 Algorithm for Surface Defect Detection in Industrial Manufacturing Field \cite{10.1007/978-3-031-44754-9_2}}
The YOLO-IMF algorithm represents a significant milestone in the application of YOLOv8 in industrial manufacturing fields. This Improved YOLOv8 variant is engineered for surface defect detection (i.e. aluminum defects), showcasing its adaptability to critical tasks in quality control and assurance within manufacturing environments.

\begin{figure}[htbp]
\centerline{\includegraphics[width=0.7\textwidth]{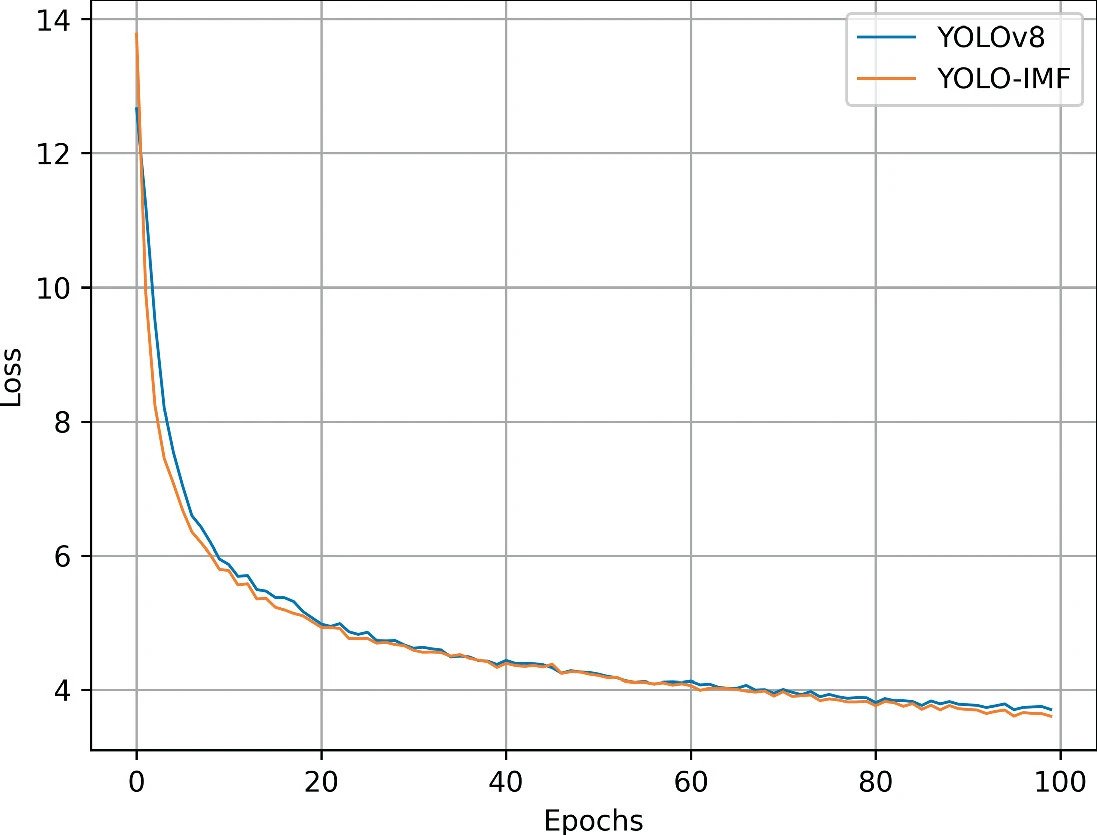}}
\caption{Loss comparison chart during training.}
\label{fig::yoloimf}
\end{figure}

The proposed YOLO-IMF algorithm replaced the CIOU loss function with the EIOU loss function. The EIOU loss function enhances bounding box regression by improving similarity measurement for small and irregularly shaped objects, enabling compatibility with various bounding box representations, and addressing common regression problems for increased stability and convergence in the algorithm. Figure \ref{fig::yoloimf} shows that the loss of YOLO-IMF decreases faster than that of YOLOv8. 

\subsubsection{A YOLOW Algorithm of Water-Crossing Object Detection \cite{app13158890}}
The paper uses the You Only Look Once (YOLO) Algorithm of Water-Crossing Object Detection to introduce innovative techniques that enhance the model’s ability to accurately identify floating objects on the water surface. It combines two modules - the SPDCS module and the SPPAUG module to achieve the best possible performance. To be more precise, the SPDCS module keeps all the data in the channel dimension, which improves the model's capacity to identify and recognize items that traverse water. Multiscale feature fusion is carried out via the SPPAUG module, which enhances the model's capacity for recognition and detection. To speed up the detection speed, the C2f module is also introduced from YOLOv8.

The proposed enhanced water-crossing object recognition algorithm based on the YOLOv5 model is able to overcome the issues of weak anti-interference, difficulty in recognizing small objects, and inability to handle complicated situations in traditional water-crossing object detection. The approach has a "correcting" mechanism and significantly lowers the probability of missed detection while increasing detection accuracy by adding three modules to the network model.

\subsubsection{Research on Detection Method for the Leakage of Underwater Pipeline by YOLOv3 \cite{zhao2020research}}
\vspace{5mm}

The paper addresses the issue of detecting oil leaks in underwater pipelines. The authors propose a detection method based on the YOLOv3 algorithm, tailored to identify oil spill points in underwater pipeline images. They tackle common underwater image issues such as distortion, blur, and low contrast by employing image enhancement techniques like Gauss filtering, brightness enhancement, sharpening, and histogram equalization.
The enhanced images are then used to train the YOLOv3 model. The approach streamlines the detection process by integrating object classification and localization into a single step, thereby improving efficiency. Experimental results indicate that the network can swiftly detect pipeline vulnerabilities with high accuracy and a low rate of missed detections. This research contributes to the field of underwater robotics and marine monitoring, potentially improving the response to and prevention of marine oil spills.

\begin{figure}[htbp]
\centerline{\includegraphics[width=0.5\textwidth]{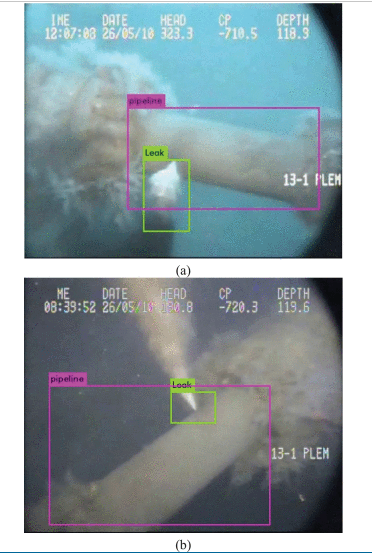}}
\caption{Pipeline with leakage points}
\label{fig}
\end{figure}

\subsubsection{Research on Oil Leakage Detection in Power Plant Oil Depot Pipeline Based on Improved YOLO v5 \cite{peng2022research}}
\vspace{5mm}

The paper discusses an enhanced YOLOv5 algorithm for detecting oil leaks in power plant pipelines, particularly in oil depots where leaks can cause significant safety incidents. The proposed improvement integrates the Convolutional Block Attention Module (CBAM) attention mechanism into YOLO v5, which helps the model to concentrate more on the characteristics of pipeline leaks and less on complex backgrounds. This integration is aimed at improving the precision of leak detection.\\
Additionally, the paper introduces Adaptively Spatial Feature Fusion, which enhances the scale invariance of features and reduces computational overhead during the inference process. The improved CBAM-YOLO v5 algorithm was tested on a self-built leak dataset from power plant pipelines, showing a 32.6\% increase in detection accuracy, a 0.76\% increase in recall rate, and a 0.76\% increase in mean Average Precision (mAP) over the original YOLO v5 model. These technical enhancements suggest that the algorithm is not only more accurate but also more efficient, making it well-suited for real-world deployment in monitoring power plant safety.

\begin{figure}[htbp]
\centerline{\includegraphics[width=0.5\textwidth]{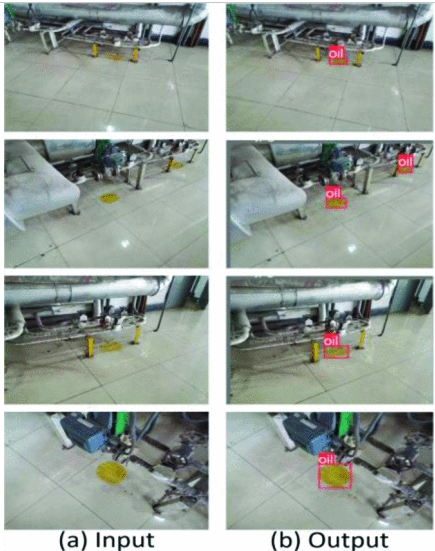}}
\caption{Detection of oil leakage based on CBAM-YOLO v5.}
\label{fig}
\end{figure}
\section{Methodology}

Our proposed methodology for detecting thermal liquid leaks utilizes the state-of-the-art YOLOv8 model and RT DETR model while incorporating several steps from data augmentation to result analysis. Initially, we collected dataset of thermal images from collaborating industry companies \cite{nleakai_dataset}, which were augmented using and Roboflow workspace to enhance the diversity and quality of the training data. This preprocessing step is critical for developing a robust model capable of handling real-world variations in leak appearances.

Upon augmentation, the dataset was uploaded to the Roboflow Workspace, where it was split into distinct sets for training, validation, and testing. The core of our methodology is the training of the YOLOv8 model and DETR model on our custom dataset with specific hyperparameters to our thermal imaging context. The model learned to identify the characteristic signatures of liquid leaks through intensive training on the prepared datasets.

To benchmark our trained models i.e. YOLOv8 and RT DETR, we compared them using existing metrices. The comparison leverages key metrics such as mean Average Precision (mAP), precision, and recall to evaluate and compare the performances of both models.

\begin{figure}[htbp]
\centerline{\includegraphics[width=1\textwidth]{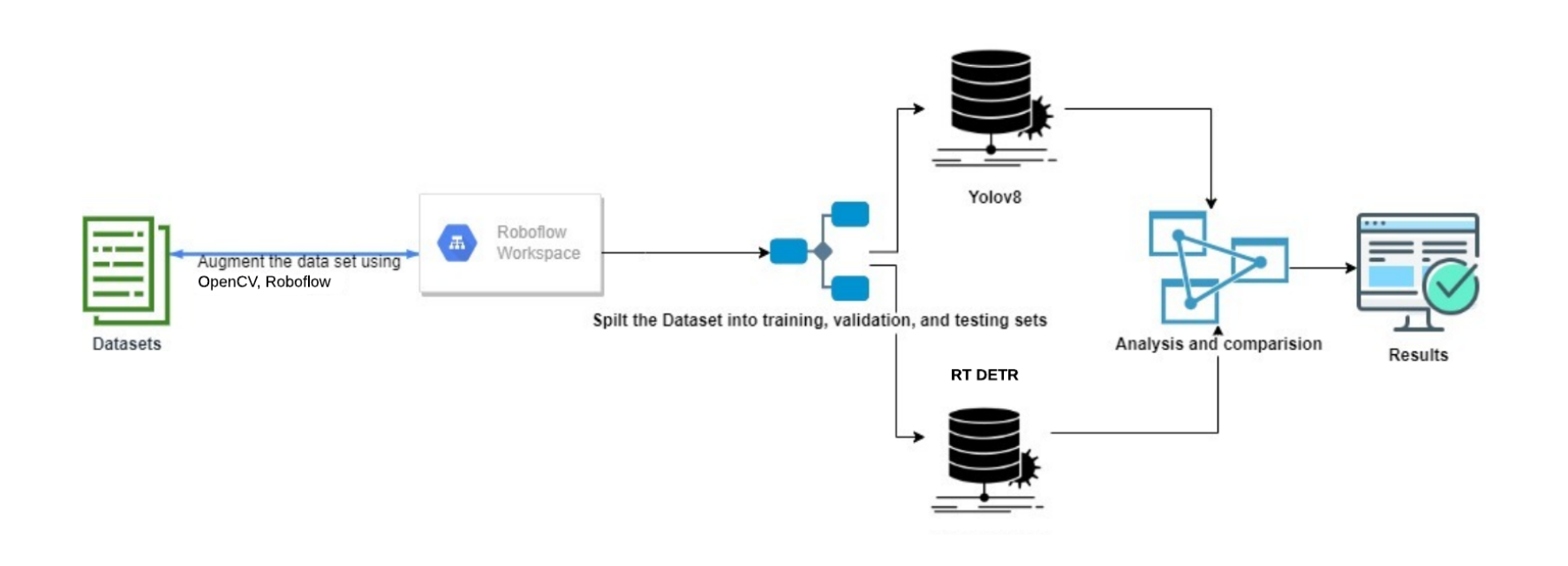}}
\caption{Methodology}
\label{fig}
\end{figure}

\subsection{Dataset}

The dataset we used for this project is a thermal dataset provided by collaborating industry companies \cite{nleakai_dataset}. The original size of the dataset before augmentation is 1,546.

We divided the original dataset into training set (80\%), validation set (10\%), and testing set (10\%). We applied multiple methods for pre-processing and augmenting the dataset. Specifically, images were reoriented to align with the desired orientation, and were resized to a standardized dimension of 640x640 pixels, ensuring uniformity in the dataset. Augmentation techniques were then applied to enhance the diversity of the dataset. Horizontal flipping was employed to create mirrored versions of the images. Additionally, color augmentation was implemented, with the hue adjusted within a range of -25° to +25°, saturation varied between -25\% and +25\%, and exposure modified within the range of -25\% to +25\%. Finally, a subtle blur of up to 2.5 pixels were applied, contributing to the overall robustness of the augmented dataset. This comprehensive series of pre-processing steps collectively aimed to improve the generalization and performance of the subsequent machine learning or computer vision models. Finally, we augmented the initial training set 80\% (1,228) using the above mentioned techniques and after data augmentation, the total size of images became 2,774. 

\subsection{YOLOv8}

Ultralytics introduced YOLOv8 as the most current edition to the YOLO family of object detection models in 2023. The name YOLO stands for "You Only Look Once," and refers to the model's ability to detect items by scanning the image through the neural network only once. YOLOv8 builds on the YOLOv5 framework with several architectural and developer experience improvements. It has a unique backbone network, a novel loss function, a cutting-edge anchor-free detecting head, and other enhancements. YOLOv8 is faster and more exact than its predecessors in the YOLO series, and it provides an all-encompassing framework for training models for image classification, object identification, and instance segmentation \cite{ultralytics}.

\begin{figure}[htbp]
\centerline{\includegraphics[width=0.5\textwidth]{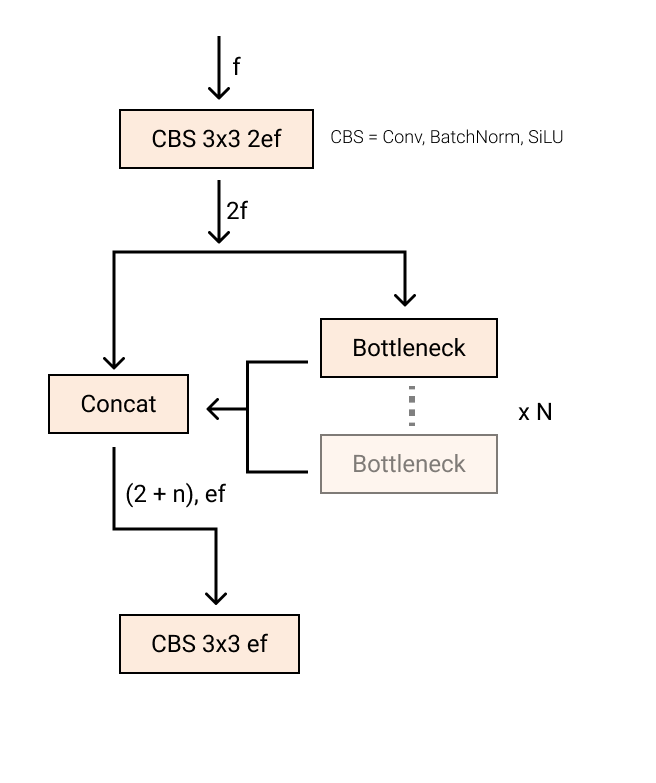}}
\caption{Improved YOLOv8 C2F Module}
\label{fig}
\end{figure}

YOLOv8 incorporates a number of architectural updates and enhancements over prior YOLO versions. The anchor-free design is one of the most significant architectural changes in YOLOv8. Anchor-based models use fixed anchor boxes to estimate the position and size of objects. Anchor-free models, on the other hand, do not use pre-defined anchor boxes and instead anticipate the location and size of objects directly, leading in more accurate object detection. Additionally, YOLOv8 has new convolutions. Figure below illustrates the modifications made to the primary building block, the replacement of C3 with C2f, and the alteration of the initial 6x6 conv in the stem with a 3x3. These modifications can improve the speed and accuracy of the mode \cite{solawetz_jan}.
A neural network's primary feature extraction component is the backbone network. From the input data, it derives useful features that help with object recognition, instance differentiation, and image classification in the future. A new backbone network for YOLOv8 has the potential to improve performance. Using the information that the backbone network retrieved, the new anchor-free detecting head predicts the location and size of objects in an image.

\begin{figure}[htbp]
\centerline{\includegraphics[width=1\textwidth]{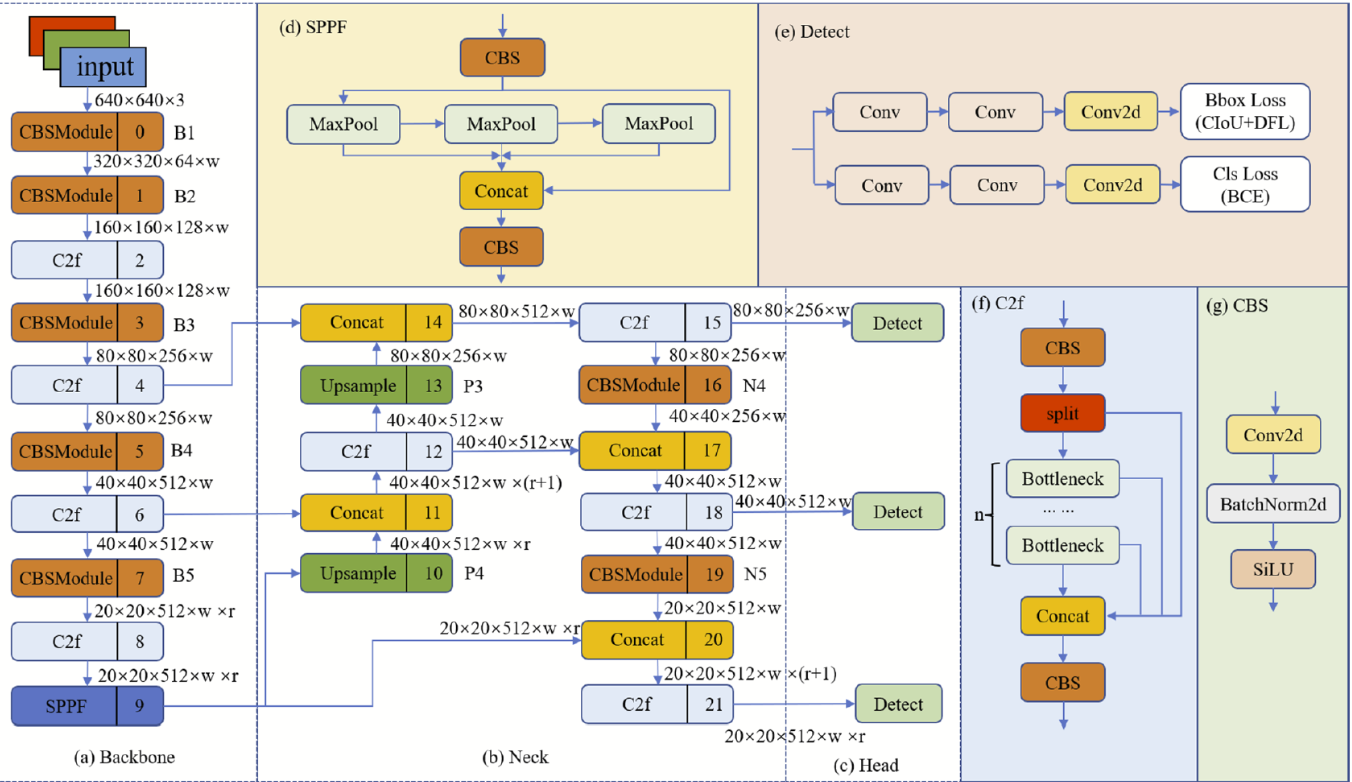}}
\caption{YOLOv8 Network}
\label{fig}
\end{figure}

In our experiment, we took advantage of YOLOv8's features, like an anchor-free architecture and an improved backbone network, to detect thermal liquid leaks. When it comes to analyzing thermal images—which present different difficulties than regular visual imagery—these traits are very helpful. YOLOv8 performed impressively in our tests, showcasing its better skills in a practical, high-stakes scenario. The model has a mean Average Precision (mAP) of 96.0 \%, indicating that it was accurate at detecting leaks. This high mAP is critical in thermal leak detection, where missing a leak can be disastrous.
Furthermore, the model exhibited a precision of 90.8\%, which signifies the reliability of the model in correctly identifying true leak instances among all detected instances. This high precision is vital in minimizing false positives, which is especially important in industrial settings to avoid unnecessary shutdowns or inspections.
The recall rate of 89.9\% reflects the model's ability to detect almost all actual leaks, a critical feature for ensuring safety and efficiency in operations where thermal liquid leaks are a concern. This high recall rate indicates that YOLOv8 is capable of detecting most leaks, thereby providing a reliable tool for early detection and prevention.
We were able to effectively tune the model for our specific use case by fine-tuning YOLOv8 with our custom dataset and altering default hyperparameters. The Adam optimizer was used to train the model across 25 epochs with an image size of 800 and a learning rate of 0.002, which contributed to its satisfactory performance.

\begin{figure}[htbp]
\centerline{\includegraphics[width=1\textwidth]{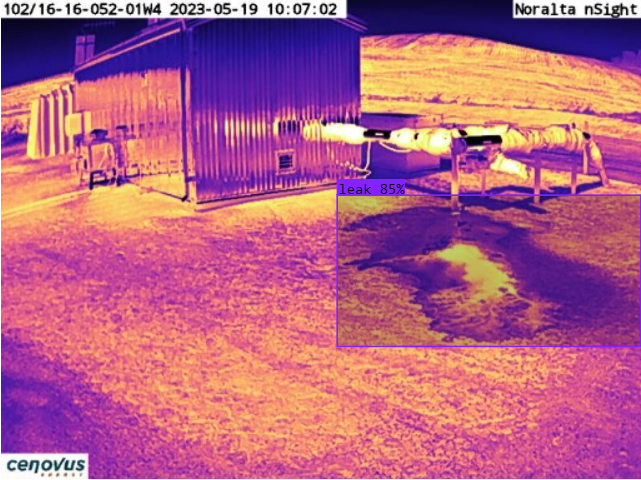}}
\caption{Sample Leak Detected}
\label{fig}
\end{figure}

In conclusion, the results from our experiment of YOLOv8 on thermal liquid leak detection confirmed its suitability and effectiveness. With an mAP of 96.0\%, a precision of 90.8\%, and a recall of 89.9\%, YOLOv8 stands out as a robust solution for detecting thermal liquid leaks, proving its potential in both theoretical and practical scenarios.

\subsection{RT-DETR}

The DETR (DEtection Transformer) model is extended by RT-DETR (Real-Time DETR), which is intended for real-time object detection in live streams and videos. By fusing the advantages of transformer topologies with a set-based global methodology, it eliminates the need for complicated post-processing procedures or anchors to enable accurate and effective object detection. By refining the model architecture and implementing strategies like feature warping to improve temporal consistency, RT-DETR attains real-time performance. It works well in situations where responsiveness in real-time and dynamic objects are important, which makes it useful for applications like autonomous systems and video surveillance. The model is fully trained, showing state-of-the-art performance in terms of speed and accuracy for real-time object detection.

\begin{figure}[htbp]
\centerline{\includegraphics[width=1\textwidth]{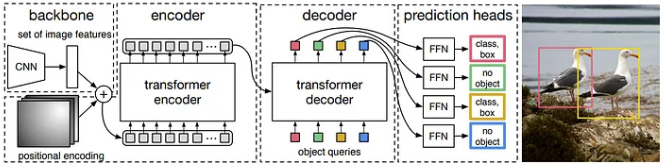}}
\caption{RT DETR Architecture}
\label{fig}
\end{figure}

The Real-Time Detection Transformer (RT DETR) model in our liquid leak detection system played a pivotal role in enhancing the efficacy of identifying leaks, particularly in scenarios involving thermal datasets. RT DETR, optimized for image classification tasks, excels in capturing intricate spatial information crucial for detecting anomalous patterns associated with liquid leaks on the ground as well as the walls in some pictures. Leveraging its transformer-based architecture, RT DETR demonstrated an exceptional ability to analyze thermal data and discern nuanced features indicative of multiple potential leaks in an image \cite{lv2023detrs}. The model's attention mechanisms facilitated the identification of subtle temperature variations and thermal anomalies, contributing to the overall accuracy of early leak detection. By harnessing the strengths of RT DETR, our system not only provided an increased sensitivity to thermal signatures associated with liquid leaks but also demonstrated its adaptability to diverse environmental, lighting conditions, and camera angles, making it a valuable asset in mitigating environmental risks and ensuring the integrity of oil operations.

\section{Results}

\begin{figure}[htbp]
    \centering
    \begin{subfigure}{0.5\textwidth}
        \centering
        \includegraphics[width=\textwidth]{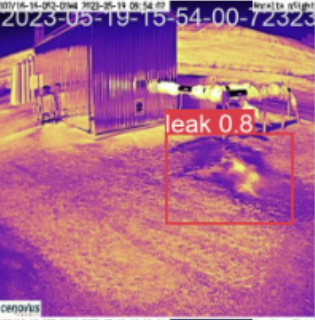}
        \caption{RT DETR sample}
        \label{fig:rtdetr}
    \end{subfigure}%
    \begin{subfigure}{0.75\textwidth}
        \centering
        \includegraphics[width=0.75\textwidth]{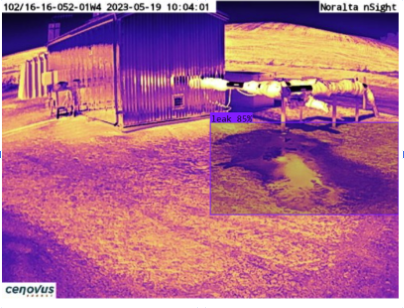}
        \caption{YOLOv8 sample}
        \label{fig:yolov8}
    \end{subfigure}
    \caption{Comparison of RT DETR and YOLOv8}
    \label{fig:comparison}
\end{figure}

\begin{table}[h]
\centering
\begin{tabular}{|c|c|c|c|}
\hline
\textbf{Model} & \textbf{Precision} & \textbf{Recall} & \textbf{mAP} \\
\hline
YOLOv8 & 90.8\% & 89.9\% & 96.0\% \\
\hline
RT-DETR & 88.1\% & 85\% & 91.7\% \\
\hline
\end{tabular}
\caption{Comparison of the two models}
\label{tab:your_table_label}
\end{table}

When evaluating the fluid leak detection system, we used two leading object detection models, YOLOv8 and RT-DETR, to evaluate their performance using key metrics such as precision, recall, and mean precision (mAP). YOLOv8 performed well in all metrics, with 90.8\% precision, 89.9\% recall, and 96.0\% mAP. On the other hand, RT-DETR showed slightly lower statistics with 88.1\% precision, 85\% recall, and 91.7\% mAP. The comparison results show that YOLOv8 outperforms RT-DETR in accurately detecting and locating fluid leaks in the system. Several factors contribute to the observed performance differences. The success of YOLOv8 is due to its efficient anchor-free approach, which can perform object detection tasks with high precision and recall. In addition, the YOLOv8 architecture is optimized for real-time applications, making it particularly effective in rapidly detecting and responding to liquid leakage events. On the other hand, although RT-DETR has demonstrated proficiency in several computer vision tasks, especially with thermal images, its set-based detection paradigm along with the resolution of the images in the dataset may not be suitable for the details of fluid leak detection systems. In conclusion, the comparison highlights YOLOv8 as the preferred choice for liquid leak detection applications and demonstrates its ability to achieve high precision, recall, and mAP. The efficiency and accuracy of the model are suitable for the real-time requirements of fluid leak detection and are proven to be superior to RT-DETR in real-world use cases.

\section{Discussion and Future Work}
\subsection{Limitations of the Work}

In acknowledging the limitations of our work, key challenges exist that may impact the overall robustness of our models. Firstly, the limitation in the diversity of our dataset and the restricted number of cameras capturing thermal images introduce a risk of overfitting, potentially hindering the models' adaptability to varied real-world conditions. A more comprehensive range of scenarios and conditions would contribute to a more resilient model. Additionally, the resolution of the images in our dataset may be insufficient, affecting the models' ability to discern subtle details crucial for accurate leak detection.

\subsection{Future Work}
We are committed to further diversifying our training dataset, potentially through the implementation of Generative Adversarial Networks (GANs) and other advanced generative models. This approach will enhance the variety of our training data, to further improve the robustness of our model against diverse conditions.

The integration of our detection system with automated response mechanisms is another possible direction for research. In addition, we anticipate conducting further studies to assess the reliability of our model in a multitude of industrial environments.

In conclusion, our research sets the stage for future advancements that will not only refine the capabilities of thermal liquid leak detection models but also contribute to safer and more efficient industrial practices worldwide.

\bibliographystyle{unsrt}
\bibliography{report}
\end{document}